\documentclass[11pt]{article}
\usepackage[T1]{fontenc}
\usepackage{amsmath,amssymb,amsthm}
\usepackage{graphicx}
\usepackage{url}
\usepackage{booktabs}
\usepackage{multirow}
\usepackage{algorithm}
\usepackage{algorithmic}
\usepackage[margin=1in]{geometry}
\usepackage{natbib}
\usepackage{xcolor}
\usepackage{float}
\usepackage[hypcap=true]{caption}
\usepackage{subcaption}
\usepackage{hyperref}
\usepackage[capitalise,noabbrev]{cleveref}

\newtheorem{theorem}{Theorem}
\newtheorem{definition}{Definition}
\newtheorem{corollary}{Corollary}

\title{Spatially-Grounded Document Retrieval via\\Patch-to-Region Relevance Propagation}

\author{Athos Georgiou\\
\textit{Independent Researcher}\\
\texttt{athrael.soju@gmail.com}
}

\date{}

\begin{document}

\maketitle

\begin{abstract}
Late-interaction multimodal retrieval models like ColPali achieve state-of-the-art document retrieval by embedding pages as images and computing fine-grained similarity between query tokens and visual patches. However, they operate at page-level granularity, limiting utility for retrieval-augmented generation (RAG) where precise context is paramount. Conversely, OCR-based systems extract structured text with bounding box coordinates but lack semantic grounding for relevance assessment. We propose a hybrid architecture that unifies these paradigms: using ColPali's patch-level similarity scores as spatial relevance filters over OCR-extracted regions. We formalize the coordinate mapping between vision transformer patch grids and OCR bounding boxes, introduce intersection metrics for relevance propagation, and establish theoretical bounds on area efficiency. We evaluate on BBox-DocVQA with ground-truth bounding boxes. For within-page localization (given correct page retrieval), ColQwen3-4B with percentile-50 thresholding achieves \textbf{59.7\% hit rate at IoU@0.5} (84.4\% at IoU@0.25, 35.8\% at IoU@0.7), with mean IoU of \textbf{0.569}, compared to $\sim$6.7\% for random region selection. Our approach reduces context tokens by \textbf{28.8\%} compared to returning all OCR regions and by \textbf{52.3\%} compared to full-page image tokens. Our approach operates at inference time without additional training. We release Snappy, an open-source implementation at \url{https://github.com/athrael-soju/Snappy}.
\end{abstract}

\section{Introduction}

Retrieval-augmented generation (RAG) has emerged as the dominant paradigm for grounding large language models in external knowledge, enabling factual responses without costly retraining. The effectiveness of RAG systems hinges on a fundamental requirement: retrieving \emph{precisely relevant} context while minimizing noise. For text corpora, this challenge is well-studied. Dense retrievers identify semantically similar passages, and chunking strategies control context granularity. However, document collections present a fundamentally harder problem.

Documents are not sequences of tokens but \emph{spatially-organized visual artifacts}. A single page may contain heterogeneous elements, including tables, figures, equations, headers, and footnotes, each carrying distinct semantic content at different spatial locations. When a user queries ``What was the Q3 revenue?'', the answer likely resides in a specific table cell, not spread across the entire page. Yet current retrieval systems operate at the wrong granularity.

Late-interaction retrievers such as ColPali \citep{faysse2025colpali} have achieved state-of-the-art performance on document retrieval benchmarks by embedding document pages directly as images. ColPali produces 1{,}024 patch embeddings (a $32 \times 32$ grid) per page, each projected to 128 dimensions. The model computes relevance through late interaction, specifically a MaxSim operation that sums the maximum similarity between each query token and all document patches. This approach elegantly sidesteps OCR errors and preserves layout semantics. However, ColPali and its variants treat pages as atomic retrieval units, making pages the finest granularity at which relevance can be assessed. For RAG applications, this is problematic: feeding a full page into a language model's context window introduces irrelevant content, increases latency, inflates costs, and, critically, dilutes the signal that the model must attend to. The retrieval system knows \emph{which page} contains the answer but not \emph{where on the page}.

Conversely, OCR-based pipelines extract text with precise bounding box coordinates, enabling structured representations of document content. Tables become rows and columns; figures receive captions; headers define hierarchy. This structural fidelity is invaluable for downstream processing. Yet OCR systems lack \emph{semantic grounding}. They cannot assess which extracted regions are relevant to a given query. A page with twenty OCR regions offers no ranking mechanism; all regions are treated as equally plausible candidates.

We observe that these paradigms are complementary. ColPali's patch-level similarity scores encode \emph{where} on a page the model attends when processing a query. This information is computed but discarded when returning page-level results. OCR systems know \emph{what} content exists and \emph{where} it is located, but not \emph{why} it matters. By unifying these signals through spatial coordinate mapping, we achieve region-level retrieval: returning only the document regions that are both structurally coherent (via OCR) and semantically relevant (via late-interaction attention).

Crucially, our approach operates at \emph{inference time} without additional training. Unlike RegionRAG \citep{li2025regionrag}, which uses a hybrid training approach combining bounding box annotations with weakly-supervised signals from unlabeled data, our method leverages ColPali's emergent patch attention as a post-hoc spatial filter. This provides flexibility: the same approach works with any OCR system providing bounding boxes and any ColPali-family model.

\subsection{Contributions}

This paper presents a hybrid architecture for spatially-grounded document retrieval:

\begin{enumerate}
    \item \textbf{Coordinate Mapping Formalism.} We formalize the mathematical correspondence between vision transformer patch grids and OCR bounding boxes, enabling spatial alignment between heterogeneous representations (\Cref{sec:coord}).
    
    \item \textbf{Relevance Propagation via Interpretability Maps.} We repurpose ColPali's late interaction mechanism to generate per-query-token similarity heatmaps, then propagate these scores to OCR regions through IoU-weighted patch-region intersection (\Cref{sec:relevance}).
    
    \item \textbf{Two-Stage Retrieval Architecture.} We introduce a two-stage architecture that enables efficient candidate retrieval before full-resolution region-level reranking (\Cref{sec:twostage}).
    
    \item \textbf{Theoretical Analysis.} We establish bounds on area efficiency as a function of patch resolution, derive expected context reduction factors, and analyze computational complexity tradeoffs (\Cref{sec:theory}).
    
    \item \textbf{Empirical Validation and Open Implementation.} We evaluate on BBox-DocVQA, demonstrating 59.7\% hit rate at IoU@0.5 with 52.3\% token savings versus full-page retrieval (\Cref{sec:eval}). We release Snappy, a complete open-source system implementing this architecture (\Cref{sec:impl}).

    \item \textbf{Model and Training Data Analysis.} We analyze how model architecture and training data affect patch-to-region localization, identifying two distinct regimes: a model-dependent regime where larger models yield substantial gains, and a precision-bound regime where patch quantization limits dominate regardless of model capacity (\Cref{sec:model_analysis}).
\end{enumerate}

On BBox-DocVQA, 59.7\% of retrieved regions achieve IoU $\geq$ 0.5 with ground-truth evidence bounding boxes, while reducing context tokens by 52.3\% compared to full-page retrieval.

\section{Background and Related Work}

\subsection{Late-Interaction Multimodal Retrieval}

\textbf{ColPali and Late Interaction.} ColPali \citep{faysse2025colpali} represents the state-of-the-art in visual document retrieval. Built on a SigLIP-So400m vision encoder, it produces 1{,}024 patch embeddings per page ($32 \times 32$ grid over $448 \times 448$ input resolution), each projected to 128 dimensions via a language model projection layer. Unlike single-vector approaches that pool visual features into a single embedding, ColPali preserves patch-level granularity and computes relevance through MaxSim, summing the maximum similarity between each query token and all document patches. This late interaction mechanism, inherited from ColBERT \citep{khattab2020colbert}, enables fine-grained matching while remaining computationally tractable for retrieval at scale.

The ViDoRe benchmark \citep{faysse2025colpali} evaluates visual document retrieval across diverse domains, measuring NDCG@5 for page-level retrieval. ColPali achieves strong performance, but the benchmark, like the model, operates at page granularity; region-level retrieval remains unexplored.

\textbf{ColPali-Family Models.} The ColPali architecture has spawned a family of late-interaction visual retrievers sharing the core patch-embedding approach. ColQwen3-4B \citep{huang2025colqwen3} combines the ColPali framework with a Qwen3-based language model, achieving state-of-the-art performance on ViDoRe while maintaining the patch-level embeddings essential to our approach. At the efficiency frontier, ColModernVBERT \citep{teiletche2025modernvbert} is a 250M-parameter variant achieving performance within 0.6 points of ColPali on NDCG@5 with 10$\times$ fewer parameters. Our approach applies to any model in this family, as all preserve the patch-level similarity structure we exploit for region-level retrieval. We evaluate both ColQwen3-4B and ColModernVBERT to demonstrate this generality and explore the accuracy-efficiency tradeoff.

\subsection{Layout-Aware Document Understanding}

The LayoutLM family \citep{xu2020layoutlm, xu2021layoutlmv2, huang2022layoutlmv3} pioneered joint modeling of text, layout, and visual features for document understanding. LayoutLMv3 introduced Word-Patch Alignment (WPA) pre-training, which predicts whether image patches corresponding to text words are masked. DocFormer \citep{appalaraju2021docformer} extends this direction with multi-modal self-attention that correlates text and visual features spatially. While conceptually related to our patch-OCR alignment, these approaches operate at \emph{pre-training time} to improve representations, whereas our approach uses patch similarities at \emph{inference time} for retrieval filtering. Critically, these models are designed for document \emph{understanding} tasks (NER, classification) rather than retrieval: they lack late interaction mechanisms and query-conditioned relevance scoring.

OCR-free approaches including Donut \citep{kim2022donut} and Pix2Struct \citep{lee2023pix2struct} perform document understanding directly from pixels. UDoP \citep{tang2023udop} unifies vision, text, and layout modalities through a generative pre-training objective, learning joint representations that could theoretically support retrieval; however, UDoP is optimized for understanding tasks and lacks the late-interaction scoring mechanism essential for efficient large-scale retrieval. LayoutReader \citep{wang2023layoutreader} addresses reading order detection, which affects OCR region quality upstream of our approach. These models excel at understanding but do not address the retrieval problem we target.

\subsection{Region-Level Document Retrieval}

\textbf{RegionRAG.} The closest existing work is RegionRAG \citep{li2025regionrag}, which shifts retrieval from document-level to semantic region-level granularity. RegionRAG clusters salient patches via BFS to produce visual region crops, requiring hybrid supervision (bounding box annotations plus pseudo-labels from unlabeled data) and a dual-objective contrastive loss. Our approach differs in three respects: (1) we operate at \emph{inference time} without additional training, (2) we output text with spatial coordinates rather than image crops, enabling direct use by text-only LLMs, and (3) OCR-derived regions align with semantic document structure (paragraphs, tables, captions) rather than visual patch connectivity. As of December 2025, RegionRAG has no public implementation or model weights available; we therefore compare against the ColPali-family baselines that both approaches share, which represents a limitation of the current comparison landscape rather than a methodological choice.

\textbf{DocVLM.} DocVLM \citep{shpigel2024docvlm} integrates OCR into vision-language models by compressing OCR features into learned queries. This represents the \emph{opposite direction} of our approach. DocVLM adds OCR to enhance VLM understanding, whereas we use late-interaction patch embeddings to filter and score OCR output for retrieval. The positioning of our approach relative to these methods is summarized in \Cref{tab:comparison}.

\begin{table}[ht]
\centering
\caption{Comparison with related approaches. Our approach is unique in achieving region-level retrieval at inference time without additional training, by propagating late-interaction patch similarities to OCR bounding boxes.}
\label{tab:comparison}
\begin{tabular}{lccc}
\toprule
\textbf{Method} & \textbf{Granularity} & \textbf{OCR Required} & \textbf{Training} \\
\midrule
ColPali & Page-level & No & Pre-trained \\
LayoutLM & Understanding & Yes & Pre-trained \\
RegionRAG & Region-level & Yes & Hybrid supervision \\
DocVLM & Understanding & Yes & Fine-tuning \\
\textbf{Ours} & \textbf{Region-level} & \textbf{Yes} & \textbf{Inference-time only} \\
\bottomrule
\end{tabular}
\end{table}

\section{Method}

\subsection{Problem Formulation}

Given a query $q$ and a document corpus $\mathcal{D}$ where each document $d \in \mathcal{D}$ consists of one or more pages, conventional visual document retrieval returns a ranked list of pages. We reformulate the problem as \emph{region-level retrieval}: return a ranked list of (page, region) pairs where each region corresponds to a semantically coherent text block (paragraph, table, figure caption, etc.) extracted via OCR.

Let $\mathcal{P} = \{p_1, \ldots, p_N\}$ denote the set of $N$ pages in the corpus, where each page $p_i$ has an associated set of OCR regions $\mathcal{R}(p_i) = \{r_1, \ldots, r_m\}$. Each region $r_j$ is characterized by its bounding box $B(r_j) = (x_1, y_1, x_2, y_2)$ in pixel coordinates and its text content $T(r_j)$. Our goal is to compute a relevance score $\text{rel}(q, r_j)$ for each region that captures both semantic relevance and spatial grounding.

\subsection{Coordinate Mapping: Patches to Bounding Boxes}
\label{sec:coord}

ColPali-family models (including ColQwen3 and ColModernVBERT evaluated in \Cref{sec:eval}) encode each page as a grid of $G \times G$ patch embeddings ($G = 32$) over an input image of resolution $I \times I$ ($I = 448$). Each patch corresponds to an $s \times s$ pixel region where $s = I/G = 14$ pixels. Patches are indexed in raster scan order (left-to-right, top-to-bottom).

\begin{definition}[Patch Coordinate Mapping]
For patch index $k \in \{0, \ldots, G^2 - 1\}$, the corresponding bounding box in pixel coordinates is:
\begin{equation}
\text{patch\_bbox}(k) = (\text{col} \cdot s, \text{row} \cdot s, (\text{col} + 1) \cdot s, (\text{row} + 1) \cdot s)
\end{equation}
where $\text{row} = \lfloor k / G \rfloor$ and $\text{col} = k \mod G$.
\end{definition}

When the original document page has resolution $(W, H)$ different from the model's input resolution $I \times I$, OCR bounding boxes must be scaled to the model's coordinate space:
\begin{equation}
B'(r) = \left(x_1 \cdot \frac{I}{W}, y_1 \cdot \frac{I}{H}, x_2 \cdot \frac{I}{W}, y_2 \cdot \frac{I}{H}\right)
\end{equation}

\subsection{Relevance Propagation via Patch Similarity}
\label{sec:relevance}

Given a query $q$ tokenized into $n$ tokens with embeddings $\{q_1, \ldots, q_n\}$ and a page with patch embeddings $\{d_1, \ldots, d_m\}$ ($m = G^2 = 1{,}024$), we compute the similarity matrix:
\begin{equation}
S \in \mathbb{R}^{n \times m} \quad \text{where} \quad S_{ij} = \text{sim}(q_i, d_j)
\end{equation}
where $\text{sim}(\cdot, \cdot)$ is cosine similarity. Standard ColPali aggregates this into a page score via MaxSim:
\begin{equation}
\text{Score}_{\text{page}}(q, p) = \sum_i \max_j S_{ij}
\end{equation}

We instead extract the spatial distribution of relevance by computing a per-patch score:
\begin{equation}
\text{score}_{\text{patch}}(j) = \max_i S_{ij}
\end{equation}
This captures the maximum relevance of patch $j$ to any query token, forming a spatial heatmap over the page.

\begin{definition}[Region Relevance Score]
For OCR region $r$ with scaled bounding box $B'(r)$, we propagate patch scores via normalized IoU-weighted aggregation:
\begin{equation}
\text{rel}(q, r) = \frac{\sum_j \text{IoU}(B'(r), \text{patch\_bbox}(j)) \cdot \text{score}_{\text{patch}}(j)}{\sum_j \text{IoU}(B'(r), \text{patch\_bbox}(j))}
\end{equation}
where the sums are over all patches $j$ with non-zero intersection. This weights each patch's contribution by its spatial overlap with the region, while normalizing by total IoU to ensure comparable scores across regions of different sizes.
\end{definition}

\subsection{Two-Stage Retrieval Architecture}
\label{sec:twostage}

Computing full patch-level similarity for all pages in a large corpus is prohibitively expensive. We introduce a two-stage architecture that balances efficiency and precision.

\textbf{Stage 1: Candidate Retrieval.} We mean-pool the $G^2$ patch embeddings to obtain a single $d$-dimensional page-level representation for efficient approximate nearest neighbor search, retrieving top-$K$ candidate pages. This stage uses standard dense retrieval techniques; our contribution focuses on Stage 2.

\textbf{Stage 2: Region Reranking.} For each candidate page, we compute full patch-level similarity and propagate scores to OCR regions as described in \Cref{sec:relevance}. Regions are ranked by their relevance scores, and top-$k$ regions are returned.

\subsection{Aggregation Strategies}

We consider alternative aggregation strategies for propagating patch scores to regions. Let $\text{covered}(r)$ denote the set of patch indices with non-zero intersection with region $r$:
\begin{equation}
\text{covered}(r) = \{j : \text{IoU}(B'(r), \text{patch\_bbox}(j)) > 0\}
\end{equation}

\textbf{Max Aggregation:}
\begin{equation}
\text{rel}_{\text{max}}(q, r) = \max_{j \in \text{covered}(r)} \text{score}_{\text{patch}}(j)
\end{equation}

\textbf{Mean Aggregation:}
\begin{equation}
\text{rel}_{\text{mean}}(q, r) = \frac{1}{|\text{covered}(r)|} \sum_{j \in \text{covered}(r)} \text{score}_{\text{patch}}(j)
\end{equation}

\textbf{IoU-Weighted (Default):} Uses the region relevance score from Definition 2:
\begin{equation}
\text{rel}_{\text{IoU}}(q, r) = \frac{\sum_j \text{IoU}(B'(r), \text{patch\_bbox}(j)) \cdot \text{score}_{\text{patch}}(j)}{\sum_j \text{IoU}(B'(r), \text{patch\_bbox}(j))}
\end{equation}

This normalizes by total IoU to produce comparable scores across regions of different sizes. Without normalization, larger regions accumulate higher scores simply by covering more patches.

The choice of aggregation strategy affects retrieval quality depending on region size and content density; we use IoU-weighted aggregation as the default based on its principled handling of partial patch overlaps while maintaining size-invariant comparability.

\section{Theoretical Analysis}
\label{sec:theory}

\subsection{Area Efficiency Bounds}

The spatial efficiency of our approach is fundamentally bounded by patch resolution. We formalize this tradeoff.

\begin{theorem}[Area Efficiency Bound]
For an OCR region with bounding box of width $w$ and height $h$ (in model coordinates), and patch size $s$, the maximum achievable area efficiency (ratio of target region area to total patch area covering it) is:
\begin{equation}
\text{efficiency} \leq \frac{w \cdot h}{(w + s) \cdot (h + s)}
\end{equation}
\end{theorem}

\begin{proof}
Consider a region with bounding box $B$ of dimensions $w \times h$. The set of patches intersecting $B$ depends on $B$'s alignment with the patch grid. In the worst case, $B$ is positioned such that it intersects partial patches on all four edges. Let the region's top-left corner fall at position $(x, y)$ within a patch. The region then spans from patch column $\lfloor x/s \rfloor$ to $\lfloor (x+w)/s \rfloor$ and from patch row $\lfloor y/s \rfloor$ to $\lfloor (y+h)/s \rfloor$. The number of intersecting patches is at most $\lceil w/s + 1 \rceil \cdot \lceil h/s + 1 \rceil$. The total area covered by these patches is at most $(w + s) \cdot (h + s)$, achieved when the region is maximally misaligned with patch boundaries. The efficiency (ratio of target area to retrieved patch area) is thus bounded by $(w \cdot h) / ((w + s) \cdot (h + s))$.
\end{proof}

\begin{corollary}
For ColPali with $s = 14$ pixels at $448 \times 448$ resolution:
\begin{itemize}
    \item A typical paragraph region ($200 \times 50$ pixels): efficiency $\leq 73\%$
    \item A table cell ($100 \times 30$ pixels): efficiency $\leq 60\%$
    \item A small label ($50 \times 20$ pixels): efficiency $\leq 46\%$
\end{itemize}
\end{corollary}

This analysis reveals that smaller regions suffer disproportionately from patch quantization. Applications requiring fine-grained localization (e.g., form field extraction) may benefit from higher-resolution patch grids or multi-scale approaches.

\subsection{Computational Complexity}

Let $N$ = number of pages, $M$ = average OCR regions per page, $G^2$ = patches per page, $n$ = query tokens, $d$ = embedding dimension.

\textbf{Page-level retrieval (baseline):} $O(N \cdot n \cdot G^2 \cdot d)$ for full MaxSim over all pages.

\textbf{Our two-stage approach:}
\begin{itemize}
    \item Stage 1: $O(N \cdot d)$ for ANN search with pooled embeddings
    \item Stage 2: $O(K \cdot n \cdot G^2 \cdot d + K \cdot M \cdot G^2)$ for full similarity on $K$ candidates plus region scoring
\end{itemize}

For $K \ll N$, the two-stage approach provides substantial speedup. With typical values ($N = 100{,}000$ pages, $K = 100$, $G = 32$, $n = 20$, $d = 128$, $M = 15$), Stage 1 reduces the search space by 1000$\times$ before the more expensive region-level computation.

\subsection{Expected Performance Bounds}
\label{sec:performance}

\subsubsection{Context Reduction}

Let $A_p$ denote the total page area and $A_r$ the area of the relevant region containing the answer. For a page with $M$ OCR regions of average area $\bar{A}$, we have $A_p \approx M \cdot \bar{A}$.

\begin{theorem}[Context Reduction Bound]
\label{thm:context}
Let $k$ be the number of top-scoring regions returned by our hybrid approach. The expected context reduction factor relative to page-level retrieval is:
\begin{equation}
\text{CRF} = \frac{A_p}{\sum_{i=1}^{k} A_{r_i}} \geq \frac{M}{k}
\end{equation}
with equality when all regions have equal area.
\end{theorem}

\begin{proof}
Page-level retrieval returns context proportional to $A_p$. Our approach returns context proportional to the total area of the top-$k$ regions, $\sum_{i=1}^k A_{r_i}$. Since each region has area at most $A_p$ and there are $M$ regions total, selecting $k$ regions yields area at most $k \cdot \bar{A} = k \cdot A_p / M$. The ratio $A_p / (k \cdot A_p / M) = M/k$ provides the lower bound.
\end{proof}

\begin{corollary}[Token Savings]
\label{cor:tokens}
For typical document parameters ($M = 15$ regions per page, $k = 3$ returned regions), the expected context reduction factor is at least $5\times$. This translates directly to proportional reductions in:
\begin{itemize}
    \item LLM inference cost (tokens processed)
    \item Response latency (context length)
    \item Attention dilution (irrelevant content in context window)
\end{itemize}
\end{corollary}

\subsubsection{Precision and Signal-to-Noise Improvement}

Score-based region ranking improves retrieval precision over random selection whenever patch scores correlate positively with relevance, an assumption validated empirically in \Cref{sec:eval}. This precision improvement directly translates to better signal-to-noise ratio in retrieved context: selecting fewer, higher-relevance regions reduces the irrelevant content that downstream LLMs must process. The cost-quality tradeoff in \Cref{tab:combined} illustrates this directly: our hybrid approach achieves the context efficiency of OCR-based selection while providing the relevance ranking that OCR alone cannot offer.

\begin{table}[ht]
\centering
\caption{Combined efficiency-quality comparison. The hybrid approach uniquely achieves both low context cost and high precision. Precision values for our approach are empirically measured (IoU@0.5 on BBox-DocVQA). $^*$$M \approx 15$ average regions per page.}
\label{tab:combined}
\begin{tabular}{lccc}
\toprule
\textbf{Method} & \textbf{Context Cost} & \textbf{Precision (IoU@0.5)} & \textbf{Best Use Case} \\
\midrule
ColPali (page-level) & High (1.0$\times$) & N/A (page-level) & Page identification \\
OCR + Random & Low (0.2$\times$) & $\sim$6.7\% (1/$M$ regions)$^*$ & Baseline \\
\textbf{Hybrid (ours)} & \textbf{Low (0.2$\times$)} & \textbf{59.7\%} & \textbf{Precise RAG} \\
\bottomrule
\end{tabular}
\end{table}

\section{Implementation: Snappy System}
\label{sec:impl}

We implement our approach in Snappy, an open-source document retrieval system available at \url{https://github.com/athrael-soju/Snappy}.

\subsection{Architecture Overview}

Snappy implements a two-stage retrieval pipeline over PDFs. Each page is rendered as an image, embedded via ColPali to produce patch-level multivectors, and optionally processed by layout-aware OCR to extract text regions with bounding boxes. Both patch embeddings and OCR metadata are stored in a vector database for retrieval.

At query time, the system: (1) encodes the query via ColPali's text encoder, (2) retrieves top-$K$ candidate pages via ANN search on pooled embeddings, (3) computes full patch-level similarity for candidates, (4) propagates scores to OCR regions via IoU-weighted aggregation, and (5) returns ranked regions with text content and bounding boxes.

\begin{figure}[ht]
\centering
\includegraphics[width=1\textwidth]{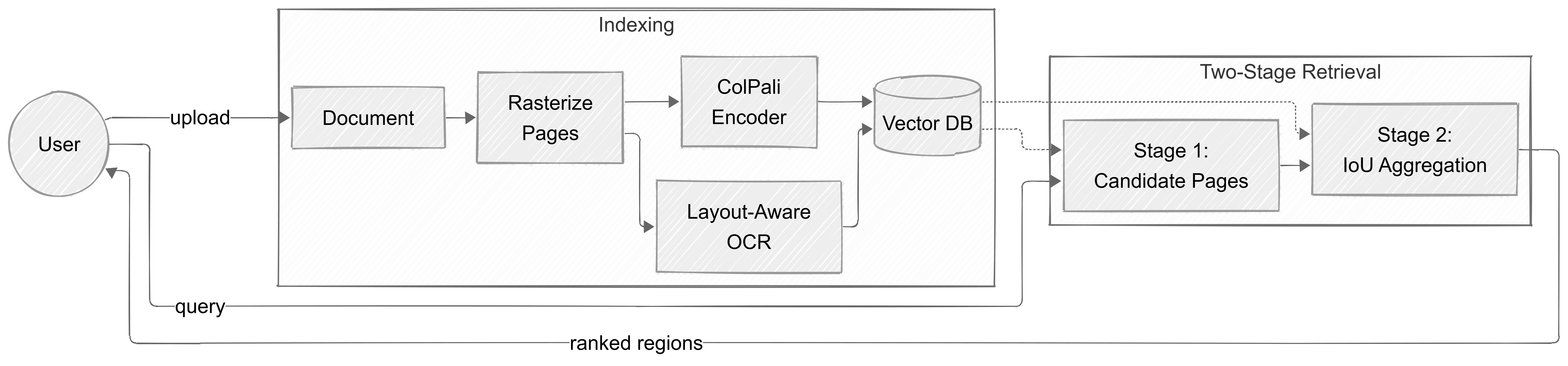}
\caption{Snappy system architecture. The indexing pipeline processes documents through parallel OCR and embedding branches. The query pipeline retrieves candidates via ANN search, computes patch-level similarities, and filters OCR regions by relevance.}
\label{fig:architecture}
\end{figure}

Unlike standard ColPali deployments that return only page-level scores, Snappy extracts the full $(n \times G^2)$ similarity matrix for downstream region scoring. The vector database stores both page-level pooled embeddings (Stage 1) and full patch multivectors (Stage 2). The overall architecture is illustrated in \Cref{fig:architecture}.

\section{Empirical Evaluation}
\label{sec:eval}

We evaluate our approach on BBox-DocVQA, measuring spatial grounding accuracy and token efficiency. Our experiments address three questions: (1) How accurately does patch-to-region relevance propagation localize relevant content? (2) How does performance vary across document categories? (3) What token savings does the approach achieve?

\textbf{Evaluation Scope.} Our evaluation targets the core contribution: \emph{region-level localization within a page}. Given a page containing the answer, can our approach identify the specific region? This complements page-level retrieval benchmarks like ViDoRe \citep{faysse2025colpali}, which measure whether the correct page is retrieved. We use IoU-based spatial grounding metrics to directly measure localization accuracy against ground-truth evidence bounding boxes. We evaluate within-page localization (Stage 2); page-level retrieval accuracy (Stage 1) is measured by existing ViDoRe benchmarks and is outside our scope.

\subsection{Experimental Setup}

\textbf{Dataset.} BBox-DocVQA \citep{yu2025bboxdocvqa} provides question-answer pairs across documents from arXiv categories including computer science (cs), economics (econ), electrical engineering (eess), mathematics (math), physics, quantitative biology (q-bio), quantitative finance (q-fin), and statistics (stat). Each QA pair includes ground-truth bounding boxes marking the evidence region containing the answer. We evaluate on 1{,}619 samples across eight categories (1{,}623 total samples minus 4 that consistently failed during OCR processing and are excluded from all reported metrics).

\textbf{Models.} We evaluate three ColPali-family models: ColQwen3-8B (8B parameters) and ColQwen3-4B (4B parameters) representing state-of-the-art accuracy at different scales, and ColModernVBERT (250M parameters) representing the efficiency frontier.

\textbf{Configuration.} We use DeepSeek-OCR with visual grounding in markdown mode, selected for its layout-aware region segmentation that groups semantically coherent blocks (paragraphs, table cells, captions) rather than individual words or lines. For region scoring, we apply 50th-percentile thresholding (denoted P50) with max token aggregation and max region scoring. The P50 threshold represents a principled midpoint validated by our ablation study (\Cref{sec:ablation}): 25th-percentile thresholds (P25) are unnecessarily inclusive (all regions selected), while 75th-percentile thresholds (P75) require careful tuning of minimum patch overlap to avoid accuracy degradation.

\textbf{Metrics.} For predicted region $B_p$ and ground-truth bounding box $B_g$, we compute:
\begin{equation}
\text{IoU}(B_p, B_g) = \frac{|B_p \cap B_g|}{|B_p \cup B_g|}
\end{equation}
We report:
\begin{itemize}
    \item \textbf{Mean IoU}: Average IoU between selected regions and ground-truth bounding boxes
    \item \textbf{Hit Rate@$\tau$}: Fraction of samples where IoU $\geq \tau$, for $\tau \in \{0.25, 0.5, 0.7\}$
    \item \textbf{Token Savings}: Percentage reduction in context tokens, computed as $(T_{\text{baseline}} - T_{\text{method}}) / T_{\text{baseline}}$, compared to (a) all OCR regions and (b) full image tokens
\end{itemize}

\subsection{Main Results}

\begin{table}[ht]
\centering
\caption{Spatial grounding accuracy on BBox-DocVQA. Hit rates indicate the fraction of samples achieving IoU at or above the specified threshold.}
\label{tab:main_results}
\begin{tabular}{lccccc}
\toprule
\textbf{Model} & \textbf{N} & \textbf{Mean IoU} & \textbf{IoU@0.25} & \textbf{IoU@0.5} & \textbf{IoU@0.7} \\
\midrule
ColModernVBERT (P50) & 1619 & 0.480 & 72.2\% & 45.5\% & 27.9\% \\
ColQwen3-4B (P50) & 1619 & 0.569 & 84.4\% & 59.7\% & 35.8\% \\
ColQwen3-8B (P50) & 1619 & 0.569 & 83.9\% & 59.8\% & 36.1\% \\
\bottomrule
\end{tabular}
\end{table}

\begin{figure}[ht]
\centering
\includegraphics[width=0.8\textwidth]{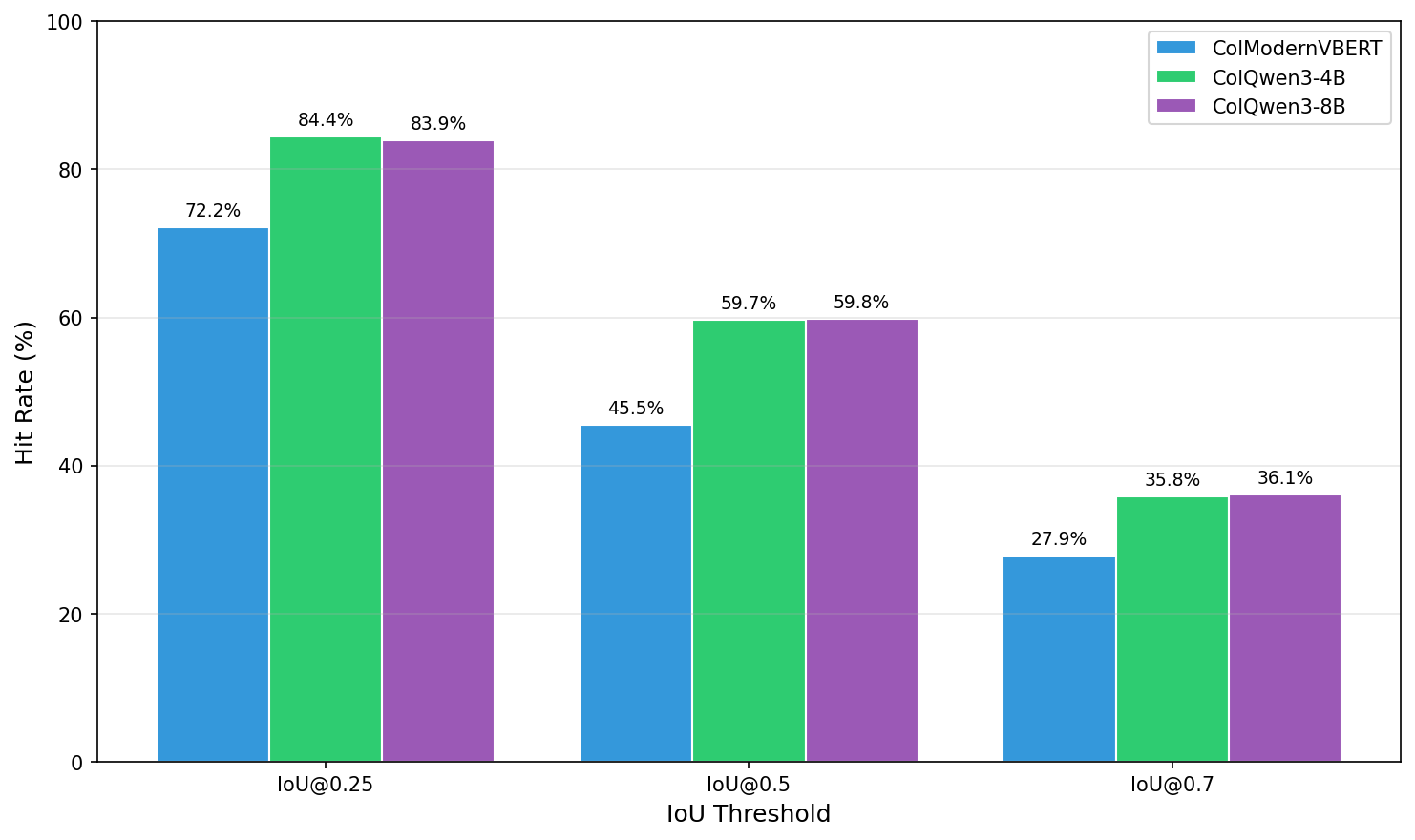}
\caption{Hit rate comparison across IoU thresholds. ColQwen3-8B and ColQwen3-4B achieve nearly identical performance, while both consistently outperform ColModernVBERT, with the gap widening at stricter thresholds.}
\label{fig:iou_thresholds}
\end{figure}

ColPali's patch attention localizes relevant regions without additional training (\Cref{tab:main_results}). ColModernVBERT (250M parameters) achieves 45.5\% hit rate at IoU@0.5 with the highest token efficiency (38.0\% savings vs OCR, 58.5\% vs full image). Scaling to ColQwen3-4B improves localization substantially (59.7\% at IoU@0.5), but further scaling to ColQwen3-8B yields negligible gains (59.8\%), demonstrating diminishing returns from parameter scaling within the same architecture (\Cref{fig:iou_thresholds}).

\subsection{Category Analysis}

\begin{table}[ht]
\centering
\caption{Performance by document category across all three models. ColQwen3-8B and ColQwen3-4B achieve nearly identical accuracy, while ColModernVBERT shows consistent degradation. Mathematics and economics documents show notably lower accuracy across all models, likely due to denser tabular content and smaller region sizes where patch quantization effects are more pronounced.}
\label{tab:category_results}
\small
\begin{tabular}{lr cc cc cc}
\toprule
& & \multicolumn{2}{c}{\textbf{ColModernVBERT}} & \multicolumn{2}{c}{\textbf{ColQwen3-4B}} & \multicolumn{2}{c}{\textbf{ColQwen3-8B}} \\
\cmidrule(lr){3-4} \cmidrule(lr){5-6} \cmidrule(lr){7-8}
\textbf{Category} & \textbf{N} & \textbf{Mean IoU} & \textbf{IoU@0.5} & \textbf{Mean IoU} & \textbf{IoU@0.5} & \textbf{Mean IoU} & \textbf{IoU@0.5} \\
\midrule
cs & 216 & 0.527 & 49.5\% & 0.697 & 75.5\% & 0.689 & 74.5\% \\
eess & 196 & 0.451 & 42.9\% & 0.656 & 78.1\% & 0.660 & 79.1\% \\
q-bio & 176 & 0.575 & 59.1\% & 0.616 & 66.5\% & 0.616 & 66.5\% \\
physics & 213 & 0.521 & 53.5\% & 0.579 & 63.8\% & 0.590 & 65.7\% \\
stat & 200 & 0.510 & 50.0\% & 0.559 & 57.0\% & 0.559 & 57.0\% \\
q-fin & 216 & 0.459 & 45.8\% & 0.543 & 59.7\% & 0.545 & 60.2\% \\
econ & 214 & 0.427 & 36.0\% & 0.513 & 46.7\% & 0.509 & 46.3\% \\
math & 188 & 0.370 & 27.1\% & 0.382 & 28.7\% & 0.378 & 27.7\% \\
\midrule
\textbf{Overall} & \textbf{1619} & \textbf{0.480} & \textbf{45.5\%} & \textbf{0.569} & \textbf{59.7\%} & \textbf{0.569} & \textbf{59.8\%} \\
\bottomrule
\end{tabular}
\end{table}

Performance varies substantially across document categories (\Cref{tab:category_results}). Computer science and electrical engineering documents achieve substantially higher localization accuracy with ColQwen3-4B (Mean IoU $\geq$ 0.65) than economics (0.513) and mathematics (0.382) documents, with physics (0.579) and quantitative biology (0.616) falling in between. ColModernVBERT shows a similar pattern but with uniformly lower scores; notably, quantitative biology achieves the highest accuracy (0.575) for the smaller model, while computer science drops from 0.697 (ColQwen3-4B) to 0.527 (ColModernVBERT). This performance gap likely reflects differences in layout complexity: economics papers frequently contain dense tables with many small cells, while mathematics papers feature equations and formulas that may span multiple regions. These smaller regions suffer disproportionately from patch quantization effects, consistent with the theoretical area efficiency bounds in \Cref{sec:theory}.

\subsection{Token Efficiency}

\begin{table}[ht]
\centering
\caption{Token efficiency comparison. Our hybrid approach with P50 filtering achieves substantial savings over both returning all OCR regions and using full image tokens.}
\label{tab:token_savings}
\begin{tabular}{lrrr}
\toprule
\textbf{Method} & \textbf{Total Tokens} & \textbf{vs All OCR} & \textbf{vs Full Image} \\
\midrule
Full Image (baseline) & 4,003,039 & -- & -- \\
All OCR Regions & 2,678,723 & -- & 33.1\% \\
ColModernVBERT (P50) & 1,661,684 & 38.0\% & 58.5\% \\
\textbf{ColQwen3-4B (P50)} & \textbf{1,908,329} & \textbf{28.8\%} & \textbf{52.3\%} \\
ColQwen3-8B (P50) & 1,974,263 & 26.3\% & 50.7\% \\
\bottomrule
\end{tabular}
\end{table}

\begin{figure}[ht]
\centering
\includegraphics[width=0.8\textwidth]{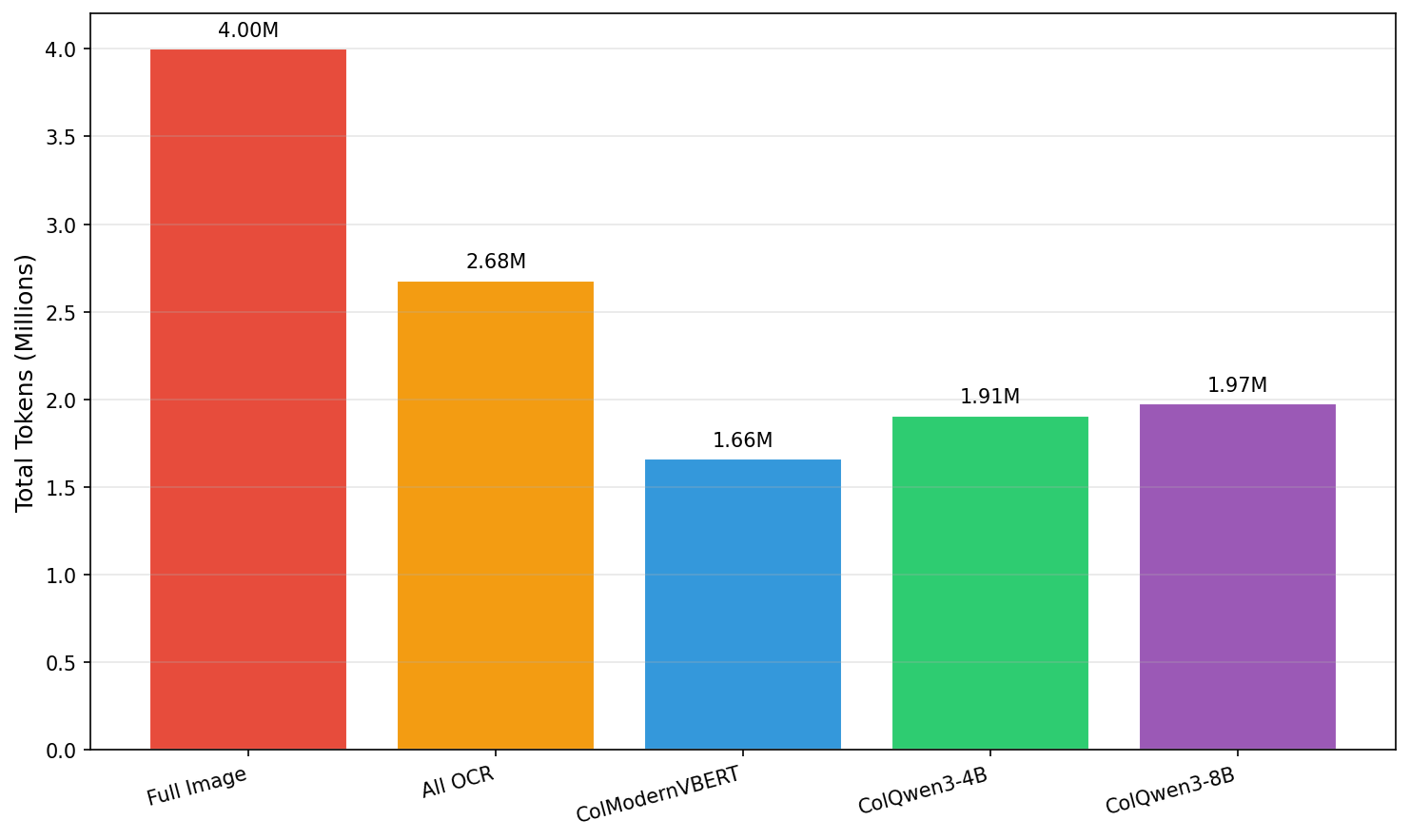}
\caption{Token usage comparison across methods. All hybrid approaches achieve substantial reductions compared to full image (4.00M) and all OCR (2.68M) baselines, with ColModernVBERT offering the greatest efficiency at 1.66M tokens. ColQwen3-8B (1.97M) and ColQwen3-4B (1.91M) achieve similar token savings.}
\label{fig:tokens}
\end{figure}

\begin{figure}[ht]
\centering
\includegraphics[width=0.6\textwidth]{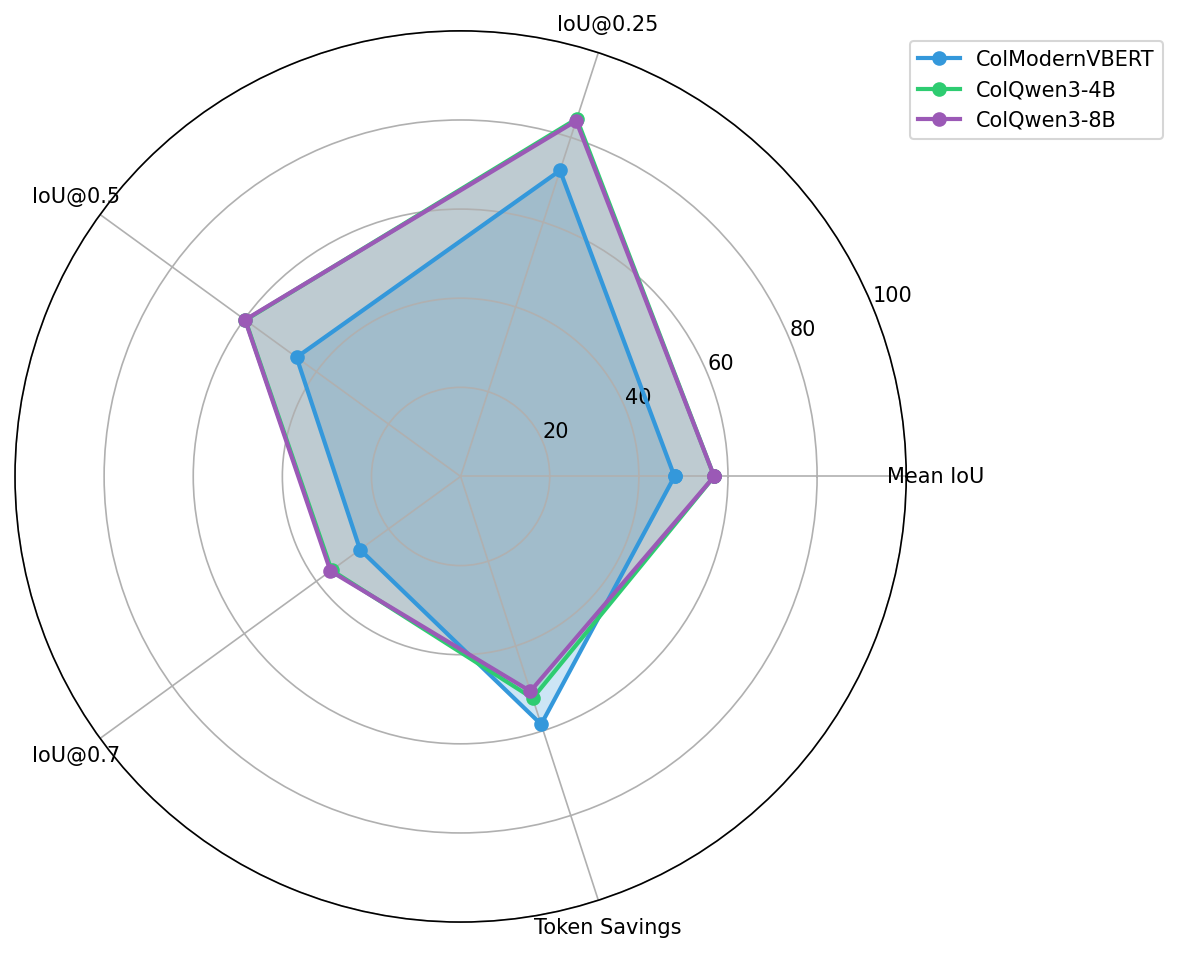}
\caption{Multi-dimensional comparison of ColQwen3-8B, ColQwen3-4B, and ColModernVBERT. ColQwen3-8B and ColQwen3-4B achieve nearly identical accuracy metrics, while ColModernVBERT offers the best token savings. Mean IoU is scaled to percentage for visual consistency.}
\label{fig:radar}
\end{figure}

We measure context tokens that would be passed to a downstream LLM. For full image retrieval, we estimate tokens using Claude's approximation: images are resized to fit within $1568 \times 1568$ pixels while maintaining aspect ratio, then token count is computed as $\lfloor (w \times h) / 750 \rfloor$. For OCR-based methods, we tokenize extracted text using tiktoken's \texttt{cl100k\_base} encoding (GPT-4 tokenizer, used as approximation). Token counts are summed across all 1{,}619 evaluation samples. The hybrid approach with P50 filtering reduces context tokens by 28.8\% compared to returning all OCR regions and by 52.3\% compared to full-page image tokens (\Cref{tab:token_savings,fig:tokens}). These savings directly reduce LLM inference costs and context window usage, consistent with the theoretical bounds in \Cref{cor:tokens}. The accuracy-efficiency tradeoff in \Cref{fig:radar} confirms that ColQwen3-8B and ColQwen3-4B achieve nearly identical accuracy, while ColModernVBERT offers the best token savings.

\subsection{Model Architecture and Training Data Effects}
\label{sec:model_analysis}

The substantial performance gap between ColQwen3-4B and ColModernVBERT (14.2 percentage points at IoU@0.5) invites analysis of how model architecture and training data affect patch-to-region relevance propagation. Notably, ColQwen3-8B achieves nearly identical performance to ColQwen3-4B (59.8\% vs 59.7\% at IoU@0.5), suggesting diminishing returns from parameter scaling within the same architecture family.

\textbf{Architectural Differences.} The models differ in attention mechanism and vision encoder design. ColQwen3-4B uses Qwen3-VL's $\sim$675M parameter vision encoder \citep{qwen2025qwen3, wang2024qwen2vl} with causal attention, while ColModernVBERT combines ModernBERT \citep{warner2024modernbert} with SigLIP2 \citep{tschannen2025siglip2} using bidirectional attention. For \emph{page-level} retrieval, bidirectional attention provides advantages (+10.6 nDCG@5 in controlled experiments), explaining ColModernVBERT's strong ViDoRe performance despite 10$\times$ fewer parameters.

\textbf{Attention Sharpness and Localization.} For \emph{region-level} localization, however, attention \emph{sharpness} matters more than global context. Localization accuracy depends on how tightly the model's attention concentrates on query-relevant patches. ColQwen3-4B's larger capacity produces sharper attention distributions, while its multi-scale features and explicit 2D spatial encoding help attention patterns respect spatial boundaries. These factors explain the 14.2 percentage point IoU@0.5 gap.

\textbf{Training Data Composition.} The models differ substantially in both pre-training scale and fine-tuning data:
\begin{itemize}
    \item \textbf{ColQwen3-4B}: Built on Qwen3-VL pre-trained on $\sim$1.4 trillion multimodal tokens \citep{qwen2025qwen3, wang2024qwen2vl} including extensive OCR data, document formats, and VQA datasets. Fine-tuned on VDR (Visual Document Retrieval), ViDoRe-ColPali-Training, and VisRAG-Ret-Train \citep{shi2024visrag}, with 63\% academic datasets (DocVQA, InfoVQA, TAT-DQA) and 37\% synthetic query-document pairs.
    \item \textbf{ColModernVBERT}: Pre-trained on $\sim$2 billion text tokens with modality alignment on The Cauldron 2 and Docmatix (21\% OCR/document content), then contrastively trained on 118k document-query pairs with 300k text-only pairs using a 2:1 text-to-image ratio.
\end{itemize}

The 700$\times$ pre-training scale difference provides ColQwen3-4B with inherently stronger document parsing and text-in-image understanding before retrieval-specific fine-tuning. However, ColModernVBERT's inclusion of text-only pairs during contrastive training provides cross-modal transfer benefits (+1.7 nDCG@5), demonstrating that the model learns text-image alignment even from pure text supervision.

\textbf{Scaling Within Architecture Family.} The near-identical performance of ColQwen3-8B and ColQwen3-4B (Mean IoU 0.569 for both; 59.8\% vs 59.7\% at IoU@0.5) reveals that doubling parameters within the same architecture provides negligible localization gains. This suggests that ColQwen3's localization accuracy is bounded by factors other than model capacity, likely patch resolution and attention pattern characteristics rather than representational capacity. The practical implication is clear: ColQwen3-4B offers the same localization quality at lower computational cost.

\textbf{Category-Specific Performance Gaps.} \Cref{tab:model_gap} reveals that the performance gap between ColQwen3-4B and ColModernVBERT varies systematically across document categories.

\begin{table}[ht]
\centering
\caption{Performance gap analysis between ColQwen3-4B and ColModernVBERT by category. ColQwen3-8B results (not shown) are within 0.01 Mean IoU of ColQwen3-4B across all categories. The gap narrows substantially for mathematics documents, suggesting both model families approach fundamental precision limits for small-region localization.}
\label{tab:model_gap}
\begin{tabular}{lcccc}
\toprule
\textbf{Category} & \textbf{ColQwen3-4B} & \textbf{ColModernVBERT} & \textbf{$\Delta$ Mean IoU} & \textbf{$\Delta$ IoU@0.5} \\
\midrule
cs & 0.697 & 0.527 & $-$0.170 & $-$26.0 pp \\
eess & 0.656 & 0.451 & $-$0.205 & $-$35.2 pp \\
q-bio & 0.616 & 0.575 & $-$0.041 & $-$7.4 pp \\
physics & 0.579 & 0.521 & $-$0.058 & $-$10.3 pp \\
stat & 0.559 & 0.510 & $-$0.049 & $-$7.0 pp \\
q-fin & 0.543 & 0.459 & $-$0.084 & $-$13.9 pp \\
econ & 0.513 & 0.427 & $-$0.086 & $-$10.7 pp \\
math & 0.382 & 0.370 & $-$0.012 & $-$1.6 pp \\
\midrule
\textbf{Overall} & \textbf{0.569} & \textbf{0.480} & $-$\textbf{0.089} & $-$\textbf{14.2 pp} \\
\bottomrule
\end{tabular}
\end{table}

\begin{figure}[ht]
\centering
\includegraphics[width=1\textwidth]{figures/iou_by_category_box.png}
\caption{IoU distribution by document category for ColQwen3-8B, ColQwen3-4B, and ColModernVBERT. Box plots show median, quartiles, and outliers. Categories ordered by mean IoU (highest to lowest). Mathematics documents show consistently low performance across all models, while computer science achieves both higher mean and greater spread.}
\label{fig:category_box}
\end{figure}

The IoU distribution across categories (\Cref{fig:category_box}) reveals that the \emph{smallest} performance gap occurs in mathematics documents ($\Delta = 0.012$ Mean IoU), where both models achieve their lowest absolute performance. This convergence suggests that mathematics documents present challenges that affect both models similarly, whether due to training data composition, difficulty parsing equations and dense notation, or other factors. The precise cause remains an open question.

Conversely, the \emph{largest} gaps appear in computer science ($\Delta = 0.170$) and electrical engineering ($\Delta = 0.205$) documents, where ColQwen3-4B's superior attention discrimination yields substantial improvements. These categories feature larger, well-separated regions (figures, code blocks, prose paragraphs) where model capacity translates directly to localization accuracy.

\textbf{Implications for Model Selection.} This analysis suggests three empirically distinct regimes:
\begin{enumerate}
    \item \textbf{Architecture-sensitive documents}: For documents with large, well-separated regions (CS, EESS, Q-Bio), ColQwen3 models substantially outperform ColModernVBERT.
    \item \textbf{Scale-insensitive within architecture}: ColQwen3-8B and ColQwen3-4B achieve nearly identical performance, indicating that doubling parameters within the same architecture yields negligible localization gains.
    \item \textbf{Uniformly challenging documents}: For mathematics and dense tabular content, all models converge to similar (lower) performance, suggesting fundamental precision limits.
\end{enumerate}

Applications should consider document characteristics when selecting models: ColModernVBERT's 16$\times$ parameter efficiency may be acceptable for mathematics-heavy corpora where all models achieve similar accuracy, while ColQwen3-4B offers the best accuracy-efficiency tradeoff for technical documentation with varied layouts. ColQwen3-8B provides no localization benefit over ColQwen3-4B despite doubled parameters.

\subsection{Error Analysis}
\label{sec:error_analysis}

To understand localization failures, we analyze the 40.3\% of samples (652/1{,}619) where the top-ranked region fails to achieve IoU $\geq$ 0.5 with the ground-truth bounding box (the complement of the 59.7\% hit rate in \Cref{tab:main_results}). Among these failures, we identify 487 samples where \emph{no} selected region achieves adequate overlap, which we analyze by root cause. The error decomposition in \Cref{tab:error_analysis} reveals two failure modes: (1) \emph{OCR ceiling failures}, where OCR segmentation did not produce any region with IoU $\geq$ 0.5 to the ground truth, and (2) \emph{selection failures}, where OCR successfully segmented the target region but our relevance propagation selected a different region.

\begin{table}[ht]
\centering
\caption{Error analysis for ColQwen3-4B failures at IoU@0.5. The dominant failure mode (85\%) is OCR segmentation ceiling: our approach cannot select regions that do not exist. Only 15\% of failures are attributable to the relevance propagation mechanism.}
\label{tab:error_analysis}
\begin{tabular}{lrrl}
\toprule
\textbf{Failure Mode} & \textbf{Count} & \textbf{\% of Failures} & \textbf{Description} \\
\midrule
OCR ceiling & 414 & 85.0\% & No OCR region overlaps GT sufficiently \\
Selection error & 73 & 15.0\% & OCR found target; wrong region selected \\
\midrule
\textbf{Total failures} & \textbf{487} & \textbf{100\%} & \\
\bottomrule
\end{tabular}
\end{table}

\textbf{OCR Ceiling Dominates.} The majority of failures (85\%) occur because the OCR system did not segment a region that adequately overlaps the ground-truth bounding box. This represents a fundamental ceiling: our relevance propagation cannot select regions that do not exist. This finding validates our architectural choice: the patch-to-region propagation mechanism works well when given appropriate candidates.

\textbf{Selection Failures by Category.} Among the 73 selection failures (where OCR found the target but we selected incorrectly), the distribution varies by category: q-bio (9.7\% selection error rate), physics and eess (6.1\%), cs (2.3\%), and econ (0.9\%). Categories with higher selection error rates may feature more semantically ambiguous layouts where multiple regions could plausibly contain the answer.

\textbf{Partial Overlap in Failures.} Even among samples that fail at IoU@0.5, 48\% achieve IoU $\geq$ 0.25, indicating substantial partial overlap. The mean IoU of failed samples is 0.250, suggesting that failures often represent near-misses rather than complete mislocalization. This has practical implications: for applications tolerant of approximate localization, the effective accuracy is higher than the IoU@0.5 metric suggests.

\textbf{Implications.} This error analysis suggests two directions for improvement: (1) better OCR segmentation to raise the ceiling, potentially through region merging or multi-scale segmentation, and (2) improved attention sharpness for the 15\% of cases where selection fails despite adequate OCR coverage. We discuss these limitations and future directions further in \Cref{sec:discussion}.

\subsection{Qualitative Examples}

\begin{figure}[ht]
\centering
\begin{subfigure}[t]{0.48\textwidth}
    \centering
    \includegraphics[width=\textwidth]{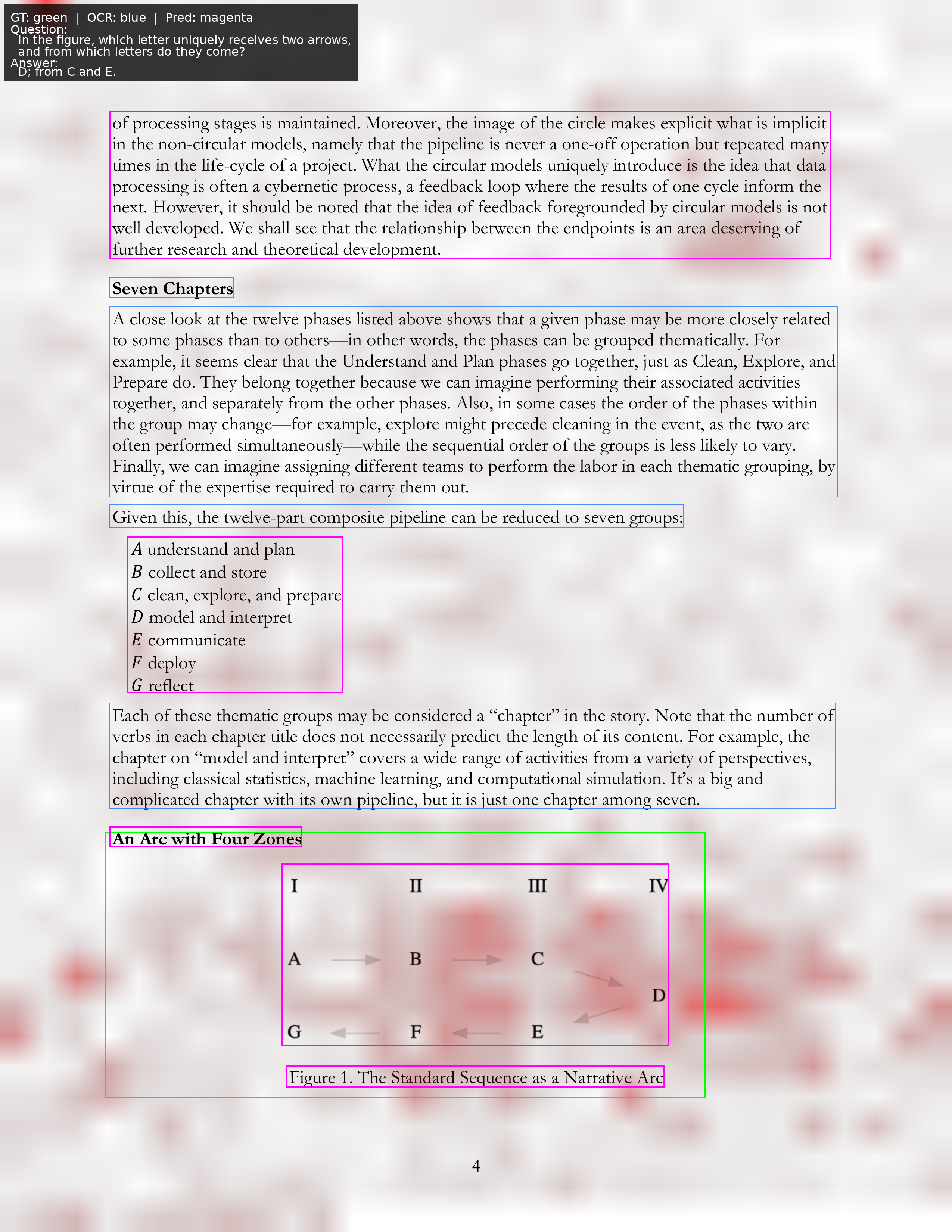}
    \caption{Query asks which letter receives two arrows in a figure. Patch attention concentrates on the diagram region; predicted box (magenta) substantially overlaps ground truth (green).}
    \label{fig:heatmap_success}
\end{subfigure}
\hfill
\begin{subfigure}[t]{0.48\textwidth}
    \centering
    \includegraphics[width=\textwidth]{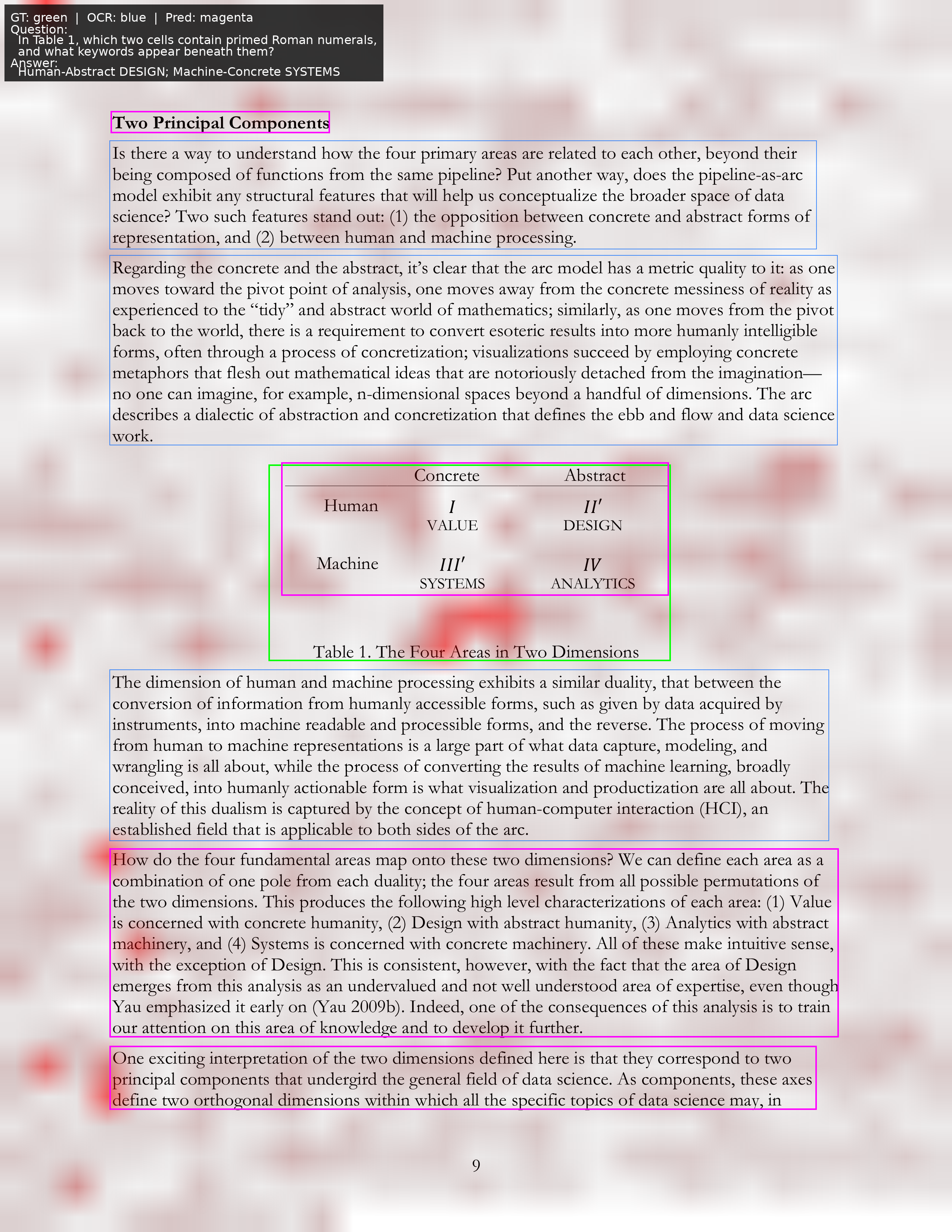}
    \caption{Query asks about table cells with Roman numerals. Attention focuses on the table structure; multiple OCR regions (blue) are scored, with the table region selected.}
    \label{fig:heatmap_table}
\end{subfigure}
\caption{Patch attention heatmaps with region overlays. Red intensity indicates patch-level similarity scores from ColQwen3-4B. Green: ground-truth bounding box. Blue: OCR-detected regions. Magenta: predicted (selected) regions. The heatmaps demonstrate that patch attention spatially concentrates on query-relevant content, enabling accurate region selection when OCR segmentation aligns with semantic boundaries.}
\label{fig:heatmaps}
\end{figure}

The patch attention heatmaps (\Cref{fig:heatmaps}) illustrate our relevance propagation mechanism on two representative samples. The heatmap overlay (red intensity) shows ColQwen3-4B's per-patch similarity scores, revealing where the model attends when processing the query. Ground-truth bounding boxes (green), OCR-detected regions (blue), and our predicted regions (magenta) demonstrate the spatial alignment between patch attention and region selection. The left example (\Cref{fig:heatmap_success}) shows a successful localization where the query references a specific figure. Patch attention strongly activates over the diagram containing the answer, and our IoU-weighted aggregation correctly selects the corresponding OCR region. The right example (\Cref{fig:heatmap_table}) demonstrates table localization, where attention distributes across the table structure and our approach selects the relevant table region despite multiple OCR candidates on the page.

These visualizations illustrate the core mechanism: ColPali's emergent patch attention, trained only for page-level retrieval, provides sufficient spatial signal to discriminate between OCR regions when propagated through our IoU-weighted scoring.

\subsection{Threshold Sensitivity Analysis}
\label{sec:ablation}

We analyze sensitivity to three hyperparameters: percentile threshold (P25, P75), region scoring method (max, weighted average), and minimum patch overlap (0.1, 0.25, 0.5). We evaluate on an 80-sample stratified subset (10 per category) across 12 configurations to enable systematic comparison.

\textbf{Parameter-Insensitive Regime at P25.} All 6 configurations using P25 thresholding produce \emph{identical} results (Mean IoU 0.605, IoU@0.5 68.8\%), regardless of region scoring or minimum overlap. This demonstrates that at sufficiently inclusive thresholds, the filtering mechanism admits nearly all regions, rendering other parameters irrelevant.

\textbf{Parameter Sensitivity at P75.} Higher percentile thresholds introduce meaningful parameter sensitivity, as shown in \Cref{tab:ablation}.

\begin{table}[ht]
\centering
\caption{Ablation study results. All P25 configurations produce identical results regardless of region scoring or minimum overlap. At P75, only minimum overlap of 0.1 causes accuracy degradation; all other P75 configurations match P25 performance.}
\label{tab:ablation}
\begin{tabular}{llcccc}
\toprule
\textbf{Region Scoring} & \textbf{Min Overlap} & \textbf{Mean IoU} & \textbf{IoU@0.25} & \textbf{IoU@0.5} & \textbf{$\Delta$ IoU@0.5} \\
\midrule
\multicolumn{6}{l}{\textit{P25 (all 6 configurations identical)}} \\
any & any & 0.605 & 91.2\% & 68.8\% & -- \\
\midrule
\multicolumn{6}{l}{\textit{P75 configurations with degraded performance}} \\
max & 0.1 & 0.566 & 86.2\% & 63.7\% & $-$5.1 pp \\
weighted\_avg & 0.1 & 0.587 & 88.8\% & 66.2\% & $-$2.6 pp \\
\midrule
\multicolumn{6}{l}{\textit{P75 with min overlap $\geq$ 0.25 (all 4 configurations identical to P25)}} \\
any & $\geq$0.25 & 0.605 & 91.2\% & 68.8\% & 0.0 pp \\
\bottomrule
\end{tabular}
\end{table}

\textbf{Minimum Overlap Dominates at P75.} The only configurations showing accuracy degradation are those with minimum patch overlap of 0.1. At this threshold, regions are scored based on patches with minimal spatial intersection, introducing noise. Increasing minimum overlap to 0.25 or 0.5 fully recovers baseline accuracy, suggesting that requiring moderate spatial correspondence between patches and regions is necessary for reliable scoring.

\textbf{Region Scoring Robustness.} At low minimum overlap (0.1), weighted average scoring (Mean IoU 0.587) outperforms max scoring (0.566), demonstrating greater robustness to noisy patch contributions. At higher overlap thresholds, both methods produce identical results, as patch-region alignment is sufficient for either aggregation strategy.

\textbf{Token Aggregation Invariance.} We additionally validated that mean and sum token aggregation produce identical per-sample IoU scores on a subset of configurations, confirming that the relative magnitude of token contributions does not affect region ranking. This suggests that patch-level attention patterns, rather than token-specific weighting, drive localization accuracy.

\textbf{Practical Implications.} These findings validate our default configuration (P50, IoU-weighted aggregation) as a robust middle ground. Lower thresholds (P25) are unnecessarily inclusive, while higher thresholds (P75) require careful tuning of minimum overlap. For practitioners, the key insight is that minimum patch overlap should be set to at least 0.25 when using stricter percentile thresholds.

\section{Discussion}
\label{sec:discussion}

\subsection{Limitations}

\textbf{Patch Resolution Bound.} As analyzed in \Cref{sec:theory}, small regions suffer from limited area efficiency due to fixed patch granularity. For ColPali's 14-pixel patches, regions smaller than approximately $35 \times 35$ pixels achieve less than 50\% area efficiency. Future work could explore multi-scale patch embeddings or dynamic resolution based on document density.

\textbf{OCR Dependency.} Region quality depends on OCR accuracy and segmentation. Poorly segmented regions (merged paragraphs, split tables) degrade retrieval quality regardless of the retrieval model's spatial attention accuracy. Layout-aware OCR systems with visual grounding mitigate this, but OCR errors propagate to retrieval. Note that OCR segmentation quality affects region boundaries but is orthogonal to our relevance scoring mechanism; our contribution is patch-to-region propagation, not OCR quality.

\textbf{Emergent Attention Limitations.} ColPali's patch attention is optimized for page-level retrieval through training, not explicitly for region-level localization. Attention may diffuse across semantically related but spatially distant regions (e.g., a header and its corresponding paragraph). We rely on OCR region boundaries to constrain this diffusion.

\textbf{Category-Specific Performance.} Empirical evaluation reveals substantial performance variation across document types. Mathematics documents achieve only 28.7\% hit rate at IoU@0.5 compared to 78.1\% for electrical engineering documents. This gap likely reflects multiple factors: training data composition favoring certain document types, difficulty parsing equations and dense mathematical notation, and potential patch quantization effects for small regions. Applications targeting specific document types should consider this variation when setting performance expectations.

\textbf{Within-Document Variance.} Variance decomposition reveals that 72\% of IoU variance is \emph{within-document} (same document, different questions) versus 28\% between documents. This suggests that question-specific semantic factors, rather than inherent document difficulty, dominate localization performance. A document may yield high IoU for one question and low IoU for another, depending on how well the query aligns with the model's learned attention patterns. This finding implies that per-category performance statistics represent averages over highly variable per-question outcomes.

\subsection{Future Directions}

\textbf{Cross-Page Region Linking.} Extending region-level retrieval to link semantically related regions across pages could enable quantitative search capabilities. For example, tracking a financial metric across quarterly reports by linking table cells containing that metric across documents.

\textbf{Elimination of Image Storage.} By storing only patch embeddings and OCR-extracted text with bounding boxes, our approach potentially eliminates the need for raw image storage after indexing. This could significantly reduce infrastructure costs for large document collections while maintaining retrieval capability.

\textbf{Region-Aware Fine-Tuning.} While our approach operates without additional training, fine-tuning ColPali with region-level supervision could improve spatial attention alignment. This would sacrifice our training-free advantage but potentially yield higher localization accuracy.

\textbf{Legal Document Retrieval and Citation Grounding.} In legal contexts, retrieval precision directly impacts downstream reliability. Recent empirical work demonstrates that even retrieval-augmented legal AI tools hallucinate 17--33\% of responses, with errors compounded by coarse retrieval granularity \citep{magesh2025hallucination}. Region-level retrieval with bounding box coordinates provides citation-bounded context: each retrieved region carries verifiable provenance, constraining generation to specific, locatable sources.

\section{Conclusion}

We presented a hybrid architecture for spatially-grounded document retrieval that unifies late-interaction retrieval with structured OCR extraction. By formalizing the coordinate mapping between ColPali's patch grid and OCR bounding boxes, we enable region-level retrieval without additional training. Our approach operates entirely at inference time, providing flexibility across OCR systems and late-interaction retrieval backends.

On BBox-DocVQA, we evaluate three ColPali-family models. ColQwen3-8B and ColQwen3-4B achieve nearly identical localization accuracy (59.8\% and 59.7\% hit rate at IoU@0.5, respectively), while ColModernVBERT achieves 45.5\%. All models provide substantial token savings compared to full-page retrieval (50.7--58.5\%), consistent with the efficiency gains predicted by \Cref{thm:context}.

Category analysis across eight arXiv domains reveals systematic performance variation: computer science and electrical engineering documents achieve higher localization accuracy (Mean IoU $\geq$ 0.65) than economics (0.513) and mathematics (0.382) documents, with physics (0.579--0.590) and quantitative biology (0.616) falling in between.

Analysis of model architecture and scaling effects reveals three empirically distinct regimes. First, ColQwen3-8B and ColQwen3-4B achieve nearly identical performance, indicating that parameter scaling within the same architecture provides negligible localization gains. Second, for architecture-sensitive document types (computer science, electrical engineering), ColQwen3 models substantially outperform ColModernVBERT (up to 35 percentage points at IoU@0.5). Third, for uniformly challenging document types (mathematics), all models converge to similar performance, suggesting fundamental precision limits. This finding has practical implications: ColQwen3-4B offers the best accuracy-efficiency tradeoff, while ColQwen3-8B provides no localization benefit despite doubled parameters.

The two-stage architecture balances computational efficiency with region-level granularity, making the approach practical for large-scale document collections. We release Snappy as an open-source implementation at \url{https://github.com/athrael-soju/Snappy}, demonstrating practical applicability for retrieval-augmented generation with reduced context windows and improved precision.

\section*{Acknowledgments}

The author thanks the ColPali team (ILLUIN Technology) for their foundational work on late-interaction document retrieval and for maintaining an open research ecosystem that enabled this work. We also thank TomoroAI for releasing the ColQwen3 embedding models under Apache 2.0 license, and the Qwen team (Alibaba Cloud) for the Qwen3-VL and Qwen3-Embedding foundation models. The Snappy system implementation is available at \url{https://github.com/athrael-soju/Snappy}.


\begin{thebibliography}{22}

\bibitem[Faysse et~al.(2025)]{faysse2025colpali}
Manuel Faysse, Hugues Sibille, Tony Wu, Bilel Omrani, Gautier Viaud, C{\'e}line Hudelot, and Pierre Colombo.
\newblock {ColPali}: Efficient Document Retrieval with Vision Language Models.
\newblock In \emph{The Thirteenth International Conference on Learning Representations (ICLR)}, 2025.
\newblock arXiv:2407.01449.

\bibitem[Khattab and Zaharia(2020)]{khattab2020colbert}
Omar Khattab and Matei Zaharia.
\newblock {ColBERT}: Efficient and Effective Passage Search via Contextualized Late Interaction over {BERT}.
\newblock In \emph{Proceedings of the 43rd International ACM SIGIR Conference on Research and Development in Information Retrieval}, pages 39--48, 2020.

\bibitem[Xu et~al.(2020)]{xu2020layoutlm}
Yiheng Xu, Minghao Li, Lei Cui, Shaohan Huang, Furu Wei, and Ming Zhou.
\newblock {LayoutLM}: Pre-training of Text and Layout for Document Image Understanding.
\newblock In \emph{Proceedings of the 26th ACM SIGKDD International Conference on Knowledge Discovery \& Data Mining}, pages 1192--1200, 2020.

\bibitem[Xu et~al.(2021)]{xu2021layoutlmv2}
Yang Xu, Yiheng Xu, Tengchao Lv, Lei Cui, Furu Wei, Guoxin Wang, Yijuan Lu, Dinei Florencio, Cha Zhang, Wanxiang Che, Min Zhang, and Lidong Zhou.
\newblock {LayoutLMv2}: Multi-modal Pre-training for Visually-rich Document Understanding.
\newblock In \emph{Proceedings of the 59th Annual Meeting of the Association for Computational Linguistics and the 11th International Joint Conference on Natural Language Processing}, pages 2579--2591, 2021.

\bibitem[Huang et~al.(2022)]{huang2022layoutlmv3}
Yupan Huang, Tengchao Lv, Lei Cui, Yutong Lu, and Furu Wei.
\newblock {LayoutLMv3}: Pre-training for Document {AI} with Unified Text and Image Masking.
\newblock In \emph{Proceedings of the 30th ACM International Conference on Multimedia}, pages 4083--4091, 2022.

\bibitem[Kim et~al.(2022)]{kim2022donut}
Geewook Kim, Teakgyu Hong, Moonbin Yim, JeongYeon Nam, Jinyoung Park, Jinyeong Yim, Wonseok Hwang, Sangdoo Yun, Dongyoon Han, and Seunghyun Park.
\newblock {OCR}-free Document Understanding Transformer.
\newblock In \emph{Proceedings of the European Conference on Computer Vision (ECCV)}, pages 498--517, 2022.

\bibitem[Lee et~al.(2023)]{lee2023pix2struct}
Kenton Lee, Mandar Joshi, Iulia Turc, Hexiang Hu, Fangyu Liu, Julian Eisenschlos, Urvashi Khandelwal, Peter Shaw, Ming-Wei Chang, and Kristina Toutanova.
\newblock {Pix2Struct}: Screenshot Parsing as Pretraining for Visual Language Understanding.
\newblock In \emph{Proceedings of the 40th International Conference on Machine Learning}, pages 18893--18912, 2023.

\bibitem[Tang et~al.(2023)]{tang2023udop}
Zineng Tang, Ziyi Yang, Guoxin Wang, Yuwei Fang, Yang Liu, Chenguang Zhu, Michael Zeng, Cha Zhang, and Mohit Bansal.
\newblock Unifying Vision, Text, and Layout for Universal Document Processing.
\newblock In \emph{Proceedings of the IEEE/CVF Conference on Computer Vision and Pattern Recognition}, pages 19254--19264, 2023.

\bibitem[Appalaraju et~al.(2021)]{appalaraju2021docformer}
Srikar Appalaraju, Bhavan Jasani, Bhargava Urala Kota, Yusheng Xie, and R. Manmatha.
\newblock {DocFormer}: End-to-End Transformer for Document Understanding.
\newblock In \emph{Proceedings of the IEEE/CVF International Conference on Computer Vision (ICCV)}, pages 993--1003, 2021.

\bibitem[Wang et~al.(2023)]{wang2023layoutreader}
Zilong Wang, Yiheng Xu, Lei Cui, Jingbo Shang, and Furu Wei.
\newblock {LayoutReader}: Pre-training of Text and Layout for Reading Order Detection.
\newblock In \emph{Proceedings of the 2023 Conference on Empirical Methods in Natural Language Processing}, pages 8367--8376, 2023.

\bibitem[Li et~al.(2025)]{li2025regionrag}
Yinglu Li, Zhiying Lu, Zhihang Liu, Yiwei Sun, Chuanbin Liu, and Hongtao Xie.
\newblock {RegionRAG}: Region-level Retrieval-Augmented Generation for Visually-Rich Documents.
\newblock arXiv:2510.27261, 2025.

\bibitem[Yu et~al.(2025)]{yu2025bboxdocvqa}
Wenhan Yu, Wang Chen, Guanqiang Qi, Weikang Li, Yang Li, Lei Sha, Deguo Xia, and Jizhou Huang.
\newblock {BBox-DocVQA}: A Large Scale Bounding Box Grounded Dataset for Enhancing Reasoning in Document Visual Question Answering.
\newblock arXiv:2511.15090, 2025.

\bibitem[Shpigel~Nacson et~al.(2024)]{shpigel2024docvlm}
Mor Shpigel~Nacson, Aviad Aberdam, Roy Ganz, Elad Ben~Avraham, Alona Golts, Yair Kittenplon, Shai Mazor, and Ron Litman.
\newblock {DocVLM}: Make Your {VLM} an Efficient Reader.
\newblock arXiv:2412.08746, 2024.

\bibitem[Teiletche et~al.(2025)]{teiletche2025modernvbert}
Paul Teiletche, Quentin Mac{\'e}, Max Conti, Antonio Loison, Gautier Viaud, Pierre Colombo, and Manuel Faysse.
\newblock {ModernVBERT}: Towards Smaller Visual Document Retrievers.
\newblock arXiv:2510.01149, 2025.

\bibitem[Magesh et~al.(2025)]{magesh2025hallucination}
Varun Magesh, Faiz Surani, Matthew Dahl, Mirac Suzgun, Christopher D. Manning, and Daniel E. Ho.
\newblock Hallucination-Free? {Assessing} the Reliability of Leading {AI} Legal Research Tools.
\newblock \emph{Journal of Empirical Legal Studies}, 22:216, 2025.

\bibitem[Shi et~al.(2024)]{shi2024visrag}
Shi Yu, Chaoyue Tang, Bokai Xu, Junbo Cui, Jieyu Zhang, Jiawei Gu, Ruhui Xue, Hongyu Lin, Xianpei Han, and Le Sun.
\newblock {VisRAG}: Vision-based Retrieval-augmented Generation on Multi-modality Documents.
\newblock arXiv:2410.10594, 2024.

\bibitem[Tschannen et~al.(2025)]{tschannen2025siglip2}
Michael Tschannen, Alexey Gritsenko, Xiao Wang, Muhammad Ferjad Naeem, Ibrahim Alabdulmohsin, Nikhil Parthasarathy, Talfan Evans, Lucas Beyer, Ye Xia, Basil Mustafa, Olivier H{\'e}naff, Jeremiah Harmsen, Andreas Steiner, and Xiaohua Zhai.
\newblock {SigLIP} 2: Multilingual Vision-Language Encoders with Improved Semantic Understanding, Localization, and Dense Features.
\newblock arXiv:2502.14786, 2025.

\bibitem[Qwen Team(2025)]{qwen2025qwen3}
Qwen Team.
\newblock Qwen3 Technical Report.
\newblock arXiv:2505.09388, 2025.

\bibitem[Wang et~al.(2024)]{wang2024qwen2vl}
Peng Wang, Shuai Bai, Sinan Tan, Shijie Wang, Zhihao Fan, Jinze Bai, Keqin Chen, Xuejing Liu, Jialin Wang, Wenbin Ge, Yang Fan, Kai Dang, Mengfei Du, Xuancheng Ren, Rui Men, Dayiheng Liu, Chang Zhou, Jingren Zhou, and Junyang Lin.
\newblock {Qwen2-VL}: Enhancing Vision-Language Model's Perception of the World at Any Resolution.
\newblock arXiv:2409.12191, 2024.

\bibitem[Huang and Tan(2025)]{huang2025colqwen3}
Xin Huang and Kye Min Tan.
\newblock Beyond Text: Unlocking True Multimodal, End-to-end {RAG} with Tomoro {ColQwen3}.
\newblock Tomoro.ai, 2025.
\newblock \url{https://tomoro.ai/insights/beyond-text-unlocking-true-multimodal-end-to-end-rag-with-tomoro-colqwen3}.

\bibitem[Warner et~al.(2024)]{warner2024modernbert}
Benjamin Warner, Antoine Chaffin, Benjamin Clavi{\'e}, Orion Weller, Oskar Hallstr{\"o}m, Said Taghadouini, Alexis Gallagher, Raja Biswas, Faisal Ladhak, Tom Aarsen, Nathan Cooper, Griffin Adams, Jeremy Howard, and Iacopo Poli.
\newblock Smarter, Better, Faster, Longer: A Modern Bidirectional Encoder for Fast, Memory Efficient, and Long Context Finetuning and Inference.
\newblock arXiv:2412.13663, 2024.

\end{thebibliography}
\end{document}